# Rule-Based Semantic Tagging.
# An Application Undergoing Dictionary Glosses

*Daniel Christen*
Lugano, Switzerland
*daniel.christen@lector.ch*

*Abstract*—The project presented in this article aims to formalize criteria and procedures in order to extract semantic information from parsed dictionary glosses. The actual purpose of the project is the generation of a semantic network (nearly an ontology) issued from a monolingual dictionary, through unsupervised procedures. Since the project involves rule-based Parsing, Semantic Tagging and Word Sense Disambiguation techniques, its outcomes may find an interest also beyond this immediate intent. The cooperation of both syntactic and semantic features in meaning construction are investigated, and procedures which allows a translation of syntactic dependencies in semantic relations are discussed. The procedures that rise from this project can be applied also to other text types than dictionary glosses, as they convert the output of a parsing process into a semantic representation. In addition some mechanism are sketched that may lead to a kind of *procedural semantics*, through which multiple paraphrases of an given expression can be generated. Which means that these techniques may find an application also in 'query expansion' strategies, interesting Information Retrieval, Search Engines and Question Answering Systems.

## I. INTRODUCTION

Starting this project, the aim was to build a lexical database extracted from a monolingual Italian dictionary through unsupervised (i.e. automatic) NLP procedures, in order to build a resource for a syntactic parser[1]. Since morphological, syntactic (for instance verb's argument structure) and semantic information is extracted from the same source dictionary, the result would be a coherent lexical database, within grammatical and semantic information is strictly meaning-specific related[2].

This paper focuses on the semantic side of the project, that consists in generating a semantic network using the content of dictionary glosses[3]. Although nowadays statistical approach seems to be the hegemonic framework for both syntactic parsing and semantic engineering, I want to show how, moving from a strict linguistic perspective, grammar-based procedures can transpose dictionary glosses into semantic tagged structures and establish congruent correspondences between word meanings.

The whole project went through four distinct phases. In this paper I will first briefly describe the two preliminary operations the dictionary underwent (section 2) and then focus on the semantic interpretation of its content (sections 3-7), which is the main theme of this paper. Section 8 shows some procedures used for enhancing the generated semantic net and sketches its further development, leading to some *procedural semantics*. What seems relevant to me, is that both the semantic net, generated through the procedures I will discuss in this paper, and the procedures itself constitute a resource that may find several applications in further NLP research and implementations.

## II. FROM DICTIONARY TO LEXICAL DATABASE

The electronic version of the source dictionary[4] has first been scanned by a dedicated software in order to extract and classify all kind of information given in its entries. In dictionaries some information is structured or quasi-structured (for instance morphological features), but a large part is given in a unstructured form. Completely unstructured data are: glosses (which I call also *paraphrase*[5]), examples of utterances, idioms and all kind of grammatical explications belonging to morphological or syntactic features. A very fine tuning of the dictionary scanner has been necessary to extract and to correctly classify these unstructured types of information. The result of the scanning process is a meaning-specific indexed lexical database, where each entry (i.e. word meaning) gets its morphological and syntactic information, plus indications belonging domain (medicine, zoology, etc.) and use (formal, familiar, rhetorical, etc.). Syntactic features such as sub-categorization and typical head-dependent occurrences were extracted from the example of utterances, that are meaning specific as well.

---

1  D. Christen, *Syntagma. A linguistic approach to parsing.*
2  The whole system is currently focused on Italian, but there are good reasons to think that the same process could be applied to other languages as well.
3  Previous projects and proceedings addressing the same theme are: Zampolli and Cappelli 1983; Calzolari 1984; Lesk 1986; Byrd and Calzolari 1987, Ahlswede and Evens 1988; Vossen, Meijs and Den Broeder 1989; Peters and Kilgarriff 1990; Boguraev 1991; Calzolari 1991; Vossen 1992; Dolan, Vanderwende and Richarson 1993; Pustejovsky, Anick and Bergler 1993; Bindi, Calzolari, Monachini, Pirrelli and Zampolli 1994; J. Véronis e N. Ide 1995; Fontenelle 1997; Fontenelle 2000; Lenci et al. 2000; Sierra and McNaught 2000. Reader will find recent reviews and discussions in: Boas 2009; Granger and Paquot 2012, Fontenelle 2008 and Fontenelle 2012.
4  The source dictionary I chose, with the kind permission of the authors, is the *Dizionario dell'italiano Sabatini Coletti* (DISC), because it is the only Italian monolingual dictionary which presents the argument structure (*valency*) of verbs, which is a crucial information for rule-based parsing systems.

From 53'000 Italian dictionary entries, about 105'000 meanings have been extracted, each provided with the above mentioned features. These data found an immediate application by an Italian parser that achieved best score at 2011 Evalita parsing task[6].

The second preliminary process the dictionary underwent, was a parsing process of the glosses (paraphrases). Parsing of dictionary glosses is possible, but the parser has to be adapted to these particular syntactic structures: it must accept as autonomous constituents phrases and clauses which usually does not come alone; it can also exclude "normal" sentences with a finite verb as their head. And, finally, it has to be very precise in handling with coordinate structures, where attachment ambiguities are extremely frequent.

This leads to the next phase of the operation, that is the transposition of the syntactic representation of parsed *paraphrases* in semantic information. The result of this process is a formal representation of glosses, which contains the semantic tagged lexical items and the structure of the semantic relations between these items. I call this formal representation Semantic Frame[7]. The whole of the generated Semantic Frames, linked to each other, constitute the Semantic Net.

### III. BUILDING THE SEMANTIC NET

---

[5] Paraphrase is a core notion in the Meaning-Text Model: Mel'čuk 2012, I 2: 45-78.

[6] C.Bosco and A. Mazzei (2012), *The Evalita 2011 Parsing Task: the Dependency Track*.

[7] Although being close, from some points of view, to Fillmore's notion of "Semantic Frame", used in FrameNet (Fillmore 1982 and Fillmore, Johnson and Petruck 2003), the formal structure for semantic representation employed in this project is an independent issue, coming from an autonomous research. Its relational structure results from an inductive method, rather than from the more usual deductive perspective used in formal semantics.

The semantic interpretation of a parsed gloss content consists in two main phases. In the first one, its content is treated as an autonomous entity: lexical items contained in a gloss are first semantically tagged and their syntactic relations are analyzed and provided with a semantic tag. That is, the dependency frame, which nodes are lexical items, will be converted in a Semantic Frame, within semantic tagged lexical entities become Semantic Units and syntactic head-dependent bounds become semantic relations.

In this first phase, the conceptual units that build a Frame are clearly nothing else than words, i.e. not specific meanings of those words. For instance, in the following frame (the gloss of the word *sail*):

TOKEN_OF(sail; THING)
PART_OF(sail; boat)

"THING" is a terminal tag (a *primitive*: see section 5) for "sail". The formal description ends at this basic-tag. The item "boat", instead, is a link to the respective Semantic Frame.

As words we find on the right side of semantic relations (like "boat" above) are in most cases polysemic, a second fundamental process must take place, that aims at individuating which of the different meanings of these words are congruent in the given context. I call this task a *relevance assignment task*, applied to the lexical content of glosses: an eminently WSD problem. Lack of space, the discussion on this second phase will take place in a further paper.

In the next sections I will discuss the first of these two phases: that is tagging the lexical content and the syntactic relations of word definitions.

### IV. PARAPHRASE TYPES

In this section I will concentrate on some typical configurations of dictionary glosses (paraphrases). The explanation of a word meaning goes generally through three distinct structures:

i) synonymy, that is one or more words of the same category of the lemma;

ii) category switch (ADV > ADJ; NOUN > VERB) introduced by stereotyped formulas;

iii) hypernymy, using a more generic word and adding specific *characteristicae* that distinguish the lemma sub-type or token from other entities of the same type.

Syntactic structure has a particular relevance in these context. Paraphrases show, in general, very stereotyped formulations, which tend to be category specific.

Agentive verb's meaning definitions are typically given by an autonomous infinite clause with its arguments, to which adverbial or circumstantial information about manner, aim or function of the given action is added. Verbs designating events are also described by an infinitive clause, which additional information refers generally to its causes or consequences.

Noun definitions are much more differentiated, depending on the type of referent. A first rough classification can be made on this referential base:

a) Nouns whose referents are human beings, considered by their physiological or sexual characteristics, are generally defined by an hypernym modified by adjectives or relative clauses expressing specific features (young, female, etc.).

b) Nouns which refers to persons with focus on their social role, public function or profession may be traduced by an hypernym and some discriminating modifiers, but often also by indefinite relative pronouns ("chi", *the one who*) followed by a relative clause, where the predicate contains the main information about the meaning of a lemma.

c) Nouns related to some behavior or to a *patient* role of a person in respect to an action or an event, are generally defined by a synonym or by a relative clause (often participial) which describes the given behavior or condition. They may also be introduced by formulas like *"[The one who is] affected/concerned by"*.

d) Nouns related to animals, plants or things which belong to some scientific, technical or institutional classification are defined by an (often domain-specific) hypernym and some modifiers (PP, Adj, relative clauses). More generic nouns of animated or non-animated things show the same syntactical structure and are semantically defined through hypernymy or synonymy. Abstract nouns (related to ideas, cultural currents, philosophical concepts, activities such as sports, etc.) are mostly presented in the same way. An alternative is the use of an indefinite pronoun followed by a relative clause: "Ciò che" *(The being, that)*.

e) Nouns expressing an act (in general deverbal nouns) are mostly defined by a formula such as "the act/action of" followed by the respective verb (infinitive mood) or directly by an infinitive clause which paraphrases the action.

f) Nouns whose referent is a feature, a characteristic and which often have a corresponding adjective are mostly introduced by a relative clause, like b) above ("ciò che", "che"; *the thing, that*).

g) In dictionary glosses often an overt denomination-function takes place: the lemma is presented as a name of a thing ("Denominazione di", "Nome di", *denomination of, name of*) designed by the following noun. This kind of definition can be also given in a synthetic way by the simple preposition "di" *(of)*, which implies something like *[that is denomination] of*. There are also glosses that have as their syntactic head terms like "Tipo/genere di" (*Type/genre of*), which corresponds more exactly to a *is_a* or *token_of*, i.e *subtype > type* relation.

V. ON ENTITIES TAGGING. SEMANTIC *PRIMITIVES*

Instead of adopting a deductive method, by establishing some inventory of abstract types of semantic tags for entities and relations (which could be easily done also by taking it from existing lists[8]), I preferred to proceed in an inductive way, moving from the semantical content of the paraphrases. Moreover it is not my purpose to deal here with the well known controversy about the notion of semantic primitives. Such primitive or basic semantic concepts emerge from the dictionary content itself. This empirical approach is also necessary in order to keep the internal coherence of the semantic network extracted from the dictionary. There are two criteria that lead to the individuation of the dictionary-specific primitive concepts:

i) the frequency of the occurrences of a word in the syntactic head position of the sentence that constitutes the paraphrase: the higher the frequency, the basic the conceptual content of a word;

ii) the stop-status of a given word in a sequence of hypernymys (Section 8 below).

Table 1 (annexed) reports a part of the words that have the highest number of occurrences in the syntactic head-position of the 105'000 glosses extracted from the source dictionary. TABLE 1 shows only words with a score above 200.

We notice that the words with the highest frequency as syntactic heads of glosses are those, which effectively can aspire to the role of semantic primitives, even simply from an intuitive point of view. Verbs like "fare" *(to do)*, with 1119 occurrences, where approximately the half of them has a causative sense, as they have an infinitive verb as object; "diventare" *(to get,* 403 occurrences*)* , "rendere", *(to make,* 582*)*, "dare" *(to give,* 335*)*, "privare" *(to deprive,* 397*)*, the modal use of "potere" *(can)*; nouns like "persona" *(person,* 1219 occurrences*)*, "parte" (*part,* indicating a meronymy relation, 834), "insieme" *(set,* 742*)*, "cosa" (*thing,* 310), "luogo" (*place,* 324*)*, "sostanza" (*substance,* 251*)*, as well as "strumento" and "attrezzo" *(instrument, tool,* 334*)*, are all words that can be empirically assumed as primitives,

---

[8] Miller and Johnson-Laird 1976; Johnson-Laird 1983, chap. 15; and standards given in EAGLES; EuroWordNet (Vossen et al. 1998) and ACQUILEX; FrameNet (cf. Baker, Fillmore and Cronin 2003).

constituting the limit of semantic breakdown of definitions in the context of this dictionary[9].

The infinite pronoun "ciò" *(what,* 7087 occurrences*)*, used in defining both unanimated and animated entities, and the pronoun "chi" *(who,* 1717*)* exclusively referred to human beings, have the highest frequency and are equally interpretable as primitive entities (respectively *thing* and *person*).

It is also necessary to observe that there are numbers of synonymic variants for the same basic concept, for instance ""cosa" (*thing,* 310 occorrenze) e "oggetto" (*object,* 174), ma anche "elemento (*element,* 416); or: "luogo" (l*ocation,* 324), "zona (*zone,* 109), "posto"(*place,* 67). The same can be said for verbs like "aumentare", "crescere" *(increase, grow)* and for the respective deverbal nouns "aumento", "crescita" *(increase, growth),* whose frequency should be summed up, in order to asses their quantitative consistency as semantic primitives.

Following the two criteria mentioned at the beginning of this section, a first inventory of provisional semantic tags for basic entities has been established. Since these tags are mostly canonical in semantic network domain, only a few examples will be given here.

Verbs receive in most cases the ACTION tag, which may be completed by more specific tags if the head-word (the hypernym) of the paraphrase corresponds to a primitive concept. For instance, if a verb is defined by a hypernym such as "muovere", "andare" *(to move, to go)*, the tag CHANGE and its sub-tag PLACE may be assigned. If the hypernym is "diventare", "crescere", "aumentare", "privare", "abbassare" *(to became, increase, grow, deprivee, decrease)* the tag is done by the sequence ACTION, CHANGE, QUALITY, to which other specifications can be added (PLUS/MINUS, DIMENSION, etc.) depending on the modifiers of the hypernym. For instance, if "crescere" (to grow) is defined as "aumentare in grandezza" (*increase in size)*, the whole tag will be:

TOKEN_OF(crescere; ACTION,CHANGE,QUALITY,DIMENSION,PLUS)

Another frequent tag is: EXPRESSION, added optionally with the sub-tag SPEECH-ACT, for verbs such as "esprimere", "manifestare" *(to express),* "dire" *(to say).* Related nouns like "espressione", "manifestazione" will therefore be tagged as THING, EXPRESSION.

The verb "essere" *(to be)* and a consistent group of other verbs belong to a different kind of predicates (*attributive predicates*), that will be discussed in the next section.

Nouns are generally tagged as THINGS, followed by other specifications such as PERSON, ANIMAL, VEGETAL, INSTRUMENT etc., if the head-word of the paraphrase matches with one of those more specific basic tags. Some nouns relate to an ACTIVITY, like "studio" (the study of some discipline), "esercizio", followed by a specifier which indicates the kind of activity, for instance: "esercizio del commercio" *(commercial activity).*

Another frequent sub-tag of THING is PART_OF, that can be assigned if head-words like "parte" *(part of)*, "elemento" *(element of),* "membro" *(member of)* appear in some syntactic contexts, typically followed by the preposition "di" *(of)* and a noun, which constitutes generally the holynym of the target-word.

Nouns are tagged as STATE or EVENT if the head-term is a primitive word such as "stato", "condizione", "evento", "fatto" *(state, condition, event, fact).* But they can also receive the tag QUALITY or MANNER if they are defined by head-words like: "caratteristica" *(characteristic),* "qualità" *(quality),* "proprietà" *(propriety),* "comportamento" *(behavior),* "atteggiamento" *(attitude).*

Nouns, moreover are tagged as ACTION or ACT_OF if the head-term (the hypernym) of the paraphrase is a verb or a verb-related noun. Also in this case, more specific tags may be added to the generic ACTION tag, for instance CHANGE with its sub-specifications (see the paragraph on verb tagging, above), or SPEECH-ACT, COGNITION, etc.

Adjectives have the generic tag QUALITY, which may be further specified through the lexical content of the paraphrase. For example "veloce" (*rapid*):

TOKEN_OF(veloce; QUALITY)
*'rapid' is a quality*
REFERS_TO(veloce; movimento)
*'rapid 'refers to 'movement'*
HAS_QUALITY(movimento; rapidità)
*[which] has the quality of 'rapidity'*

is completed by the following information:

HAS_TAG(movimento,ACTION,CHANGE,PLACE)
REFERS_TO(veloce,ACTION,CHANGE,PLACE)

Through hypernymy-chains (section 8) the semantic information of an entity can be completed in most cases. That is: since a verb like "correre" *(to run)* is defined by the more generic verb "muoversi" *(to move)*, which is its hypernym, modified by the adverb "velocemente" (related to the adjective "veloce", *rapid)*, the target verb "correre" can inherit the tags of its hypernym:

TOKEN_OF(correre; muovere[si])
TOKEN_OF(correre; CHANGE,PLACE)
HAS_QUALITY(correre; MANNER(veloce))

A dedicated mechanism provides to rise the attributes of the hypernym on the level of the target word (the glossed lemma).

The tagging of entities like things, events, states, actions, qualities constitutes the first process the content of paraphrases undergoes. The basic tags assigned to entities are directly involved in the second process, which addresses the tagging of the relations between the entities inside a given paraphrase. This second process will be discussed in the next section.

---
[9] See: Mel'čuk 2012, II, 4, for instance 184-188.

## VI. ON TAGGING RELATIONS

The classical representation of a semantic relation has the following schematic representation, where A and B are the meanings of the words /a/ and /b/ respectively, and the arrow designs a tagged semantic relation:

$$/a/(A) = sem.rel => /b/(B)$$

or the logical form: sem.rel(/a/(A); /b/(B))

where the semantic relation is a predicate and the two word meanings are its arguments. A graphical representation would treat A and B as *nodes* and their bound as a tagged arch. The term on the right of the semantic relation /b/(B) may assume three different instances:

i)  /b/(B) = TAG
    That is a primitive, no further analyzed term

ii) /b/(B) = MNG(B)
    That is a link to a specific meaning "B" of /b/

iii) /b/(B) = /b/
    That is an unspecified link to a word /b/

Instances i) and iii) can be reached already in the actual phase of the whole process, since words have received a terminal semantic tag during the previous process, which satisfies instance (i); or, by default, they are left unspecified, and remain a simply link to another word (not a meaning) in the network.

Instance ii) is not reached unless lexical entities in the paraphrases are not processed by a sense disambiguation module, which assigns to each word, in a given paraphrase, the link to its congruent meaning in that context.

It is needless to say that many of semantic tags used in these project do not match with some standard inventory of current semantic tagging frameworks. This comes from a need to customize semantic description in the perspective of the applications it is conceived for. In any way, since tags are conventional, the switch to a different codification is always possible. This project aims rather at formalizing the procedures which give a semantic interpretation to the syntactic dependency frames of the parsed dictionary paraphrases.

As mentioned above, the previous semantic tagging of entities plays often a central role in selecting the congruent type of semantic relation between the entities itself. For instance, a subject of a verb tagged as ACTION can be easily tagged as AGENT_OF. The same principle acts for most of the arguments included in a verb's argument structure.

Modifiers of nouns, such as adjectives and relative clauses, are interpreted as a HAS_QUALITY relation. For example:

gorilla: Grande scimmia africana, con pelle nera ricoperta da pelo grigio scuro e con piedi prensili
*(gorilla: a kind of big African ape with black skin covered of gray hair and prehensile feet)*

The gloss has the syntactic tree shown in TABLE 2 (annexed), which is transposed in the following Semantic Frame:

| |
|---|
| TOKEN_OF(gorilla,scimmia) *'gorilla' is an 'ape'* |
| HAS_QUALITY(gorilla,grande) *'gorilla' is 'big'* |
| HAS_QUALITY (gorilla,africano) *'gorilla' has the quality 'African'* |

Besides adjectives, there are also two PP modifiers introduced by the preposition "con" *(with)*. The heads of these PPs, "pelle" *(skin)* and "piedi" *(feet)*, are modified by adjectives and a relative clause.

In this context the preposition "con" expresses a HAS_PART semantic relation.

The nouns "pelle" and "piedi" are relied to their modifiers by the HAS_QUALITY relation. And the verb of the relative clause has "pelo" as AGENT and the trace related to "pelle" as OBJECT/PATIENT.

The resulting Semantic Frame of "gorilla" is shown here below:

| |
|---|
| LEMMA: "gorilla"    MNG: 41551    CAT: NOUN |
| "gorilla" =TOKEN_OF=> "scimmia" *ape* (NOUN) |
| "gorilla" =HAS_QUALITY=> "grande" *big* (ADJ) |
| "gorilla" =HAS_QUALITY=> "africano" *african* (ADJ) |
| "gorilla" =HAS_PART=> "pelle" *skin* (NOUN) |
| "gorilla" =HAS_PART=> "piede" *foot* (NOUN) |
| "pelle" =HAS_QUALITY=> "nero" *black* (ADJ) |
| "pelo" =AGNT_OF=> "ricoprire" *cover* (VERB) |
| "pelle" =OBJ_OF=> "ricoprire" *cover* (VERB) |
| "ricoprire" =HAS_AGNT=> "pelo" *hair* (NOUN) |
| "pelo" =HAS_QUALITY=> "grigio" *gray* (ADJ) |
| "grigio" =HAS_QUALITY=> "scuro" *dark* (ADJ) |
| "piede" =HAS_QUALITY=> "prensile" *prehensile* (ADJ) |

Since prepositions in most cases do not have an univocal sense, PP noun modifiers are often ambiguous. For instance "con", we have seen in the definition of "gorilla", corresponds to a meronymy (HAS_PART) relation when it introduces the modifier of a noun:

"acacia" : "Pianta arborea o arbustiva con rami spinosi"

*(tree or bush with thorny branches)*

But when it introduces the modifier of a verb, it may express also other types of relations, such as:

HAS_INSTRUMENT:
"abbacinare" : "Abbagliare qlcu. con una luce intensa"
*(to dazzle someone with an intense light)*
"abbottonare" : "Chiudere con bottoni
*(to close with buttons)*

HAS_QUALITY,HAS_MANNER:
"abbarbicare" : "Attaccarsi con tenacia a un appiglio"
*(to clench tenaciously at a hold)*
"abbaiare" : "Gridare con rabbia e insistentemente"
*(to shout with rage and insistently)*
"abbordare" : "imboccare con risolutezza
*(to start in a resolute way)*
"abbandonare" : "affidarsi a qlcu. con fiducia"
*(to trust somebody)*

ATTRIBUTION,RELATION_TO
"accompagnare" : "Unire qlco. con o ad altro"
*(to unite something/somebody with another)*
"accompagnare" : "accordarsi, armonizzarsi con qlco."
*(to match, to harmonize with something)*
"accomunare" : "Mettere in comune qlco. con qlcu."
*(to share something with somebody)*

In some cases disambiguation is possible, for instance when the dependent is provided with a terminal tag such as INSTRUMENT, QUALITY, MANNER. But often this information is lacking, since practically every kind of thing, even abstract things, can have an instrumental role in some context. The HAS_MANNER relation can instead easier be identified, since names with MANNER or QUALITY value have been detected most of the time during the previous entity tagging process.

In many cases we have to content ourselves to leave these tags in their ambiguous form, postponing the selection of a definitive tag to the next phase of the process, i.e. the relevance assignment to the lexical content of a gloss (see section 3, above).

The preposition "per" *(for)*, introducing a noun modifier can be interpreted as well as a HAS_FUNCTION -with wide sense coverage -relation, such in paraphrases like "macchina per lavare" (*washing machine*), "vestito molto elegante, per feste o ricevimenti" *(very elegant dress for party and ceremonies)*, but also as a causal relation: "qualsiasi sostanza che, per la sua durezza, può raschiare le asperità superficiali di un materiale" *(each substance that can have an abrasive effect due to its hardness)*; or even belong to some gray semantic zone between cause and manner ("abituale": "che è tale per abitudine", *which is habitual in that way*). The preposition "per" can also mean a destination (concrete or abstract) of an action, a sentiment or an attitude: "orrore per qlcu. o per qlco." *([to feel] horror towards somebody or something)*. In all these cases, the tag of the dependent, if assigned, helps in many cases the disambiguation task.

Verbs can be often semantically ambiguous in dictionary definitions as well. Verbs tagged as CHANGE may be ambiguous between the sense 'to move' (CHANGE, PLACE) and the sense 'to become' (CHANGE, QUALITY), like "salire" *(to go down)* and "scendere" *(to go up)* which may refer both to a movement in a physical space and to an increment or decrement of a quality (temperature, for instance). Only if the dependent is provided with a terminal tag the right relation can be selected.

Expressions like "caratterizzato da" *(characterized by)* may refer both to a QUALITY (color, form etc.) or to a part of an object, expressing therefore a meronymy (HAS_PART) relation. In these cases as well, the terminal tag of the dependent plays a crucial role in disambiguation.

Two interesting aspects arouse immediate attention. First: in spite of the wide variety of formulations used in the dictionary, most of the expressed semantic relations can be reduced to a finite set of basic types, which may be tagged. But often the same type of semantic relation has a multitude of different formulations, which can make the translating task quite difficult. Thus, an inventory of all the different expressions belonging to a same relation type has to be made. They constitute classes of expressions that are related by 'family resemblances' in the sense of Wittgenstein[10].

Second: a consistent part of the verbal material used in word glosses has not a relevant semantic content but a merely functional role. I call this class of expressions *attributive functors*, and I will discuss it in the next section.

VII. FUNCTORS: ATTRIBUTIVE PREDICATES

Numerous expressions employed in dictionary glosses have a merely functional role: thus not their content is considered in the Semantic Frame, but only the semantic relation they mediate.

Expressions like "caratterizzato da", "costituito/composto di..." *(characterized by; constituted of)* can be immediately traduced into a meronymy (HAS_PART) boundary. Some others, like "in maniera, in modo..." *(in a [...] manner/way)*, coupled with an adjective, denote clearly a quality or manner attribution, which relies the target verb or noun to its specific features. There is a *family* of verbs, used in noun-definitions, which all express the function of an object: "usato/impiegato per", "adibito a", "volto a", "che serve a", "di cui lo stato si serve per" *(used/employed for; addressed to; which serves to)*, which function may be indicated by an agentive verb or a noun denoting an action: "usato per tagliare", "adibito alla mietitura" *(used for cutting, employed in reaping)*. And there are corresponding families of deverbal nouns which have exactly the same attribution role, such as "uso", "impiego" *(use, employment)*. Also a causal (HAS_CAUSE) attribution to an event or to a state-of-things is done by a closed family of expression such as "causato da", "dovuto a", "conseguente a" *(caused by, due to, consequence of)*, which can be inventoried.

---

[10] L.Wittgenstein, *Philosophical Investigations*, 67.

Another class of semantical "empty" expressions are those which relate a lemma to its hypernym through a denomination ("nome di", denominazione di": *noun of, denomination of*) or through an open type assignment ("tipo di", "genere di: *type of, kind of*).

In all these cases, the verbal material that expresses the attributive function can be deleted, thus only the extracted formal relation finds its place in the Semantic Frame, as will be shown through the next example.

The short gloss below contains three occurrences of attributive predicates:

1 zool. (al pl., iniziale maiusc.) Tipo di animali, al quale appartiene anche l'uomo, caratterizzati dalla presenza di uno scheletro interno...
*[vertebrate] is a kind of animals, to which mankind belongs too, that are characterized by the presence of an internal skeleton [which axis coincide with the vertebral column or dorsal spine]*

There are three attributive predicates in this gloss:

"tipo di" (*type of:* in reality a TOKEN_OF attribution)
"caratterizzato da" (a HAS_PART attribution)
"la presenza di" (an 'existence' attribution)

During the syntax-to-semantics translation process, these predicates keep first their full lexical form. Thus the characteristics of the main word "vertebrato" are distributed on different levels, and does not immediately refer to it.

| I | ATTRIBUTION,TOKEN_OF(vertebrate, type) |
|---|---|
| II | TOKEN_OF(type, animal) |
| III | ATTRIBUTION,HAS_PART(type, characterized) |
| IV | HAS_PART(characterized, presence) |
| V | ATTRIBUTION,EXISTENCE(presence, skeleton) |

A dedicated mechanism provides to raise the content of these attributive predicates to the upper level and to delete the attributive predicate itself, step by step, proceeding bottom-up:

| I | ATTRIBUTION,TOKEN_OF(vertebrate, type) |
|---|---|
| II | TOKEN_OF(type, animal) |
| III | ATTRIBUTION,HAS_PART(type, characterized) |
| IV | HAS_PART(characterized, skeleton) |
| V | *--raising and deletion*  |

| I | ATTRIBUTION,TOKEN_OF(vertebrato,type) |
|---|---|
| II | TOKEN_OF(type,animal) |
| III | ATTRIBUTION,HAS_PART(type,skeleton) |
| IV | *-- raising*  |
| V | -- |

| I | ATTRIBUTION,TOKEN_OF(vertebrate,type) |
|---|---|
| II | TOKEN_OF(type,animal) |
| III | HAS_PART(type,skeleton) |
| IV | *-- deletion* |
| V | -- |

| I | TOKEN_OF(vertebrate, animal) |
|---|---|
| II | *-- raising and deletion*  |
| III | HAS_PART(type,skeleton) |
| IV | -- |
| V | -- |

The last step raises the proprieties of the hypernym on the level of the lemma:

| I | TOKEN_OF(vertebrate, animal) |
|---|---|
| II | HAS_PART(vertebrate,skeleton) |
| III | *-- raising*  |
| IV | -- |
| V | -- |

The resulting Semantic Frame for "vertebrate" shows as this:

```
LEMMA: "vertebrate" MNG: 103952        CAT: NOUN

101671.0 "vertebrate" TOKEN_OF 101671.2 "animal"
101671.0 "vertebrate" HAS_PART 101671.5 "skeleton"
101671.1 "vertebrate" HAS_TOKEN: 101671.15 "man"
101671.2 "animal" HAS_TAG: THING ANIMAL
101671.5 "skeleton" HAS_QUALITY: 101671.6 "internal"
101671.5 "skeleton" HAS_QUALITY: 101671.7 "coincide"
101671.6 "internal" HAS_TAG: QUALITY
101671.7 "coincide" HAS_SUBJ: 101671.8 "axis"
101671.7 "coincide" HAS_SPACE: 101671.9 "column"
101671.7 "coincide" HAS_SPACE: 101671.11 "spine"
101671.8 "axis" HAS_TAG: THING
101671.9 "column" HAS_QUALITY: 101671 "vertebral"
101671.12 "vertebral" HAS_TAG: QUALITY
101671.11 "spine" HAS_QUALITY: 101671.12 "dorsal"
101671.12 "dorsal" HAS_TAG: QUALITY
101671.15 "man" HAS_TAG: THING PERSON
```

The same mechanism interests attributive expressions of a potential feature, such as "capace di", "in grado di" *(capable*

*of, able to)*. In the example below, the feature is an agentive role (AGENT_OF), since the dependent of the attributive predicate is an infinite verb ("muovere", *move):*

"animale" :"Ogni organismo sensibile in grado di muoversi spontaneamente, compreso l'uomo"
*(animal: each organism which is able to move spontaneously, mankind included)*

Also in this case, the AGENT_OF feature is raised on the level of the lemma "animale", the attributive predicate is deleted and the absolute quantifier ("ogni", *every*) as well. Which gives the following Frame:

LEMMA: "*animal*" MNG: 4775 CAT: NOUN
"animal" =TOKEN_OF=> *organism* (NOUN)
"animal" =HAS_QUALITY=> *sensible* (ADJ)
"animal" =AGNT_OF, POTENTIAL=> *move* (VERB)
"move" =HAS_TAG=> CHANGE(PLACE)
"move" =HAS_QUALITY=> *spontaneously* (ADV)
"animal" =HAS_PART,TOKEN=> *man* (NOUN)
"man" =HAS_TAG=> PERSON()

The sub-tag POTENTIAL plays an important role in the context of word glosses like these, since it allows its interpretation within a fuzzy logic frame. Hence, even a paraplegic cat or a mule forced to walk through whipping can still be considered animals.

The same belongs to adverbs and adjectives whose function is to add a relative value to the meaning of the word they modify. Example are adjectives and adverbs like "normal","normally", "often", such as in the definition of the adjective "grande" (*big*):

:LEX "*big*"  :CAT ADJ
*big* =TOKEN_OF=> QUALITY
"$head = AGENT_OF => *exceed*
*measure* = OBJ_OF => *exceed*
*measure* = HAS_QUALITY => *normal*

## VIII. DERIVATED RELATIONS

### A. Taxonomies

Hyperymy plays an important role in word definitions, and therefore it can be used for generating taxonomies and even some kind of an ontology structure, which reflects the dictionary-specific world representation.
In this section three procedures will be showed, through which the semantic net can be enhanced.
Hypernnymy has the transitivity propriety:

token_of (x,y) ∧ token_of (y,z) → token_of (x,z)

An immediate consequence is that it allows the automatic generation of concatenated hypenymys (*hypernym chains*):

"orango";64055 =TOKEN_OF => *"scimmia";*85532
"*ape*";85532 =TOKEN_OF=> "*mammal*";55168
"*mammal*";55168 =TOKEN_OF=> "*vertebrate*";103953
"*vertebrate*";103953 =TOKEN_OF=> "*animal*";4770;
"*animal*";4770 =TOKEN_OF=> "*organism*";64296
"*animal*";4770; =TOKEN_OF=> "*beast*";11098

Since a token inherits the attributes of its type, also the TOKEN_OF relation and its content can be passed from the more general term to its token. The result are taxonomies like the one here below:

"orango";64055 =TOKEN_OF => "scimmia";85532
*orango is an ape*
"orango";64055 =TOKEN_OF=> "mammifero";55168
*orango is a mammal*
"orango";64055 =TOKEN_OF=> "vertebrato";103953
*orango is a vertebrate*
"orango";64055 =TOKEN_OF=> "organismo";64296
*orango is an organism*
"orango";64055 =TOKEN_OF=> "bestia";11098
*orango is a beast*

To aspire to real consistency, these relations must first undergo the relevance assignment process (section 3), in order to select the congruent meanings to which the TOKEN_OF relation can be applied. For instance, some animal activist would surely protest against the machine inference that associates an *orangutan with* a *beast.*

### B. Inverse relations

Another process for enhancing automatically the information content of the semantic net is to collect the content of inverse relations and to add into the Semantic Frames. This produces useful data (for example for dealing with dependencies ambiguities in parsing tasks) in particular when applied to PART_OF (i.e. meronymy/holonymy) and QUALITY_OF relations. For instance, if one converts every occurrence of the relation:

x, HAS_QUALITY, "white"

into the inverse relation:

"white", QUALITY_OF, x

the Frame of the given meaning of "white" can be supplied with a link to all entities the quality "white" belongs to in the source-dictionary. For instance, starting from "acetosella" *(shamrock)* and ending with "zibibbo" *(Muskat from Alexandria),* 295 new data have been added to the main meaning of the adjective "white", in which we find all kind of things: wines, like *chardonnay*, flowers, white animals and animals with some typical white spots, and the *opossum*, which hair shows a shade of "white". We get even the notion that:

11196 "*white*" NEG,HAS_QUALITY: 20076 "*colored*"

that is: "white" things are not colored (reader may feel free to include Chomsky's "green colorless ideas" in this class of beings or not).

*C. Tagging semantic roles*

A third example of derivative information that can enhance the semantic net concerns semantic roles. Thematic grids, assigning the congruent name to the participants of an action designed by a verb, can be automatically filled using the information given in the corresponding nouns. For instance, since the frame of the noun "acquirente" (*buyer*) mentions it as AGENT_OF the verbs "acquistare" and "comprare" (two synonyms for *to buy*)

1270.0 AGNT_OF("acquirente";"acquistare" VERB)
1270.0 AGNT_OF("acquirente"; "comprare" VERB)

and "compratore" is also an AGENT_OF "comprare"

20607.0 AGNT_OF("compratore"; "comprare" VERB)
(*buyer is agent of to buy*)

these information can be added, by inverting the relation, to the frames of those verbs:

1270.0 HAS_AGNT("acquistare" ; "acquirente" NOUN)
1270.0 HAS_AGNT("comprare"; "acquirente" NOUN)
20607.0 HAS_AGNT("comprare"; "compratore" NOUN)

Thus some participants of the action designed by the verb receive automatically their specific names. Analogously the information given in the frames "venditore" (*seller*), "venduto" (*thing that has been sold*), "negozio" (*shop, store*):

101214.0 AGNT_OF("venditore"; "vendere" VERB)
101221.0 OBJ_OF("venduto"; "vendere" VERB)
59885.0 TOKEN_OF("negozio"; "locale" NOUN)
59885.0 PLACE_OF("negozio"; "vendere" VERB)
59885.1 HAS_TAG("locale"; PLACE)

are added to the frame of "vendere" (*to sell*) as follows:

101182 HAS_AGNT("vendere"; "venditore")
101182 HAS_OBJ("vendere"; "venduto")
101182 HAS_PLACE("vendere"; "negozio";)

giving information about its participants and even the indication of a typical location where the action takes place.

I want to conclude this list of examples, showing how these information propagates through the whole net, involving also other words and meanings.
The gloss of the meaning 20148 of "commerciare" (*to trade*) relies this verb to:

20148.0 TOKEN_OF("commerciare"; "comprare" VERB)
20148.0 TOKEN_OF("commerciare"; "vendere" VERB)

And another meaning (20146) of the same verb relates this activity to the object "commercio" (*trade*):

20146.0 TOKEN_OF("commerciare"; "esercitare" VERB)
20146.0 HAS_OBJ("commerciare"; "commercio" NOUN)
20146.2 HAS_SPEC("commercio"; "prodotto" NOUN)

which is, in turn, an activity described by a frame relating it to the verbs "comprare" and "vendere" (*buy, sell*):

20149.0 TOKEN_OF(*trade*; *activity* )
20149.2 TOKEN_OF( *trade*; 20149.3 *buy*)
20149.2 TOKEN_OF(*trade*; 20149.5 *sell*)
20149.3 HAS_OBJ(*buy*; 20149.6 *product*)
20149.5 HAS_OBJ( *sell*; 20149.6 *product*)

Notice that this link affects a category switch from noun to verb, which is a fundamental mechanism towards a procedural semantic framework, where synonymous semantic contents can be identified beyond their actual morphological ad syntactic expression, and different formulations of a same content can be related to each other or even translated the one to the other[11].
In the case of the given example, the object "prodotto" *(product)* cannot be inherited by the two hypernyms of "commercio", the verbs "comprare" and "vendere" (*buy* and *sell*), because the inheritance mechanism works only unidirectional, from a type down to a token (or sub-type).
But, instead, the given meanings of "commerciare" (*to trade*) and "commercio" (*trade* Noun) can inherit the proprieties of their hypernyms. Therefore the following information can be added to their frames:

20146.0 AGENT_OF("commerciare"; ""acquirente")
20146.0 AGENT_OF("commerciare"; ""compratore")
20146.0 AGENT_OF("commerciare"; "venditore")
20146.0 OBJ_OF("commerciare"; "venduto")
20146.0 PLACE_OF("commerciare"; "negozio" NOUN)

---

[11] For a similar use of the notion of synonymy, related to the concept of paraphrase, see Mel'čuk 2012, I 2.

Now let us make a step forward, looking at the formalized gloss of "commerciante" (20145) *(merchant)*:

20145.0 TOKEN_OF("commerciante"; 20145"chi"*who*)
20145.0 HAS_SPEC("commerciante"; 20145.2 "mestiere)
20145.0 AGNT_OF("commerciante"; 20145.3 "esercitare")
20145.1 HAS_TAG("chi"; THING PERSON)
20145.2 HAS_TAG("mestiere"; ACTIVITY)
20145.3 HAS_TAG("esercitare"; ACTION)
20145.3 HAS_OBJ("esercitare"; 20145.4 "commercio")
20145.4 HAS_TAG("commercio"; THING)

That is: *a merchant is a PERSON which professional activity consists in trading*. The frame shows the AGENT_OF an activity ("esercitare", that is the same verb used in the definition of "commerciare" 20146, above) which object is "commercio" (the same of 20146).

A dedicated procedure operates the following inference:
  p & q ¬ r

p) 20145.0 AGNT_OF("commerciante"; 20145.3 "esercitare")
   20145.3 HAS_OBJ("esercitare"; 20145.4 "commercio")

q) 20146.0 TOKEN_OF("commerciare"; "esercitare")
   20146.0 HAS_OBJ("commerciare"; "commercio")

r) 20145.0 AGNT_OF("commerciante"; 20145.3 "commerciare")

Which means that "commerciante" can be added, through the inversion of the relation AGENT_OF, to the existing agents of "commerciare" 20146. To which the system can therefore automatically count also two other entities related to "commerciante":

56002 TOKEN_OF("mercante"; 56002.1 "commerciante")
56002 TOKEN_OF("mercante"; 56002.2 "negoziante")

(*a merchant is a trader*)
(*a merchant is a storekeeper*)

New agentive participant denominations are now related to these verb and nouns, which can play a crucial role in Information Retrieval tasks but also in solving ambiguities a parser has to deal constantly with.
And we get also the semantic relevant information that "commerciare" and "commercio" are activities that belongs to human beings due to the tag PERSON of "commerciante",

## IX. Conclusion

This paper aimed at describing criteria and procedures that have been developed in order to extract semantic information from a machine readable Italian dictionary, which preliminary underwent a parsing process. Both the semantic net generated and the procedures designed for these purpose constitute resources that may be integrated in further research and applied to other projects. The theoretical framework and the strategies from which this project moves are strictly grammar-based, and therefore a deep investigation of meaning structures is implied, considering the strength interaction between syntax and semantics. Semantic relations between etymological and morphological heterogeneous words have been detected through procedures which can be applied also to other text types than dictionary glosses. Moreover, some mechanisms that emerged from this research suggest a perspective of semantic analysis that goes beyond the morphological and syntactic material of the surface level of expression. Since in a dictionary words are defined by other words, and words are mostly polysemic, a second phase of this project is already ongoing, addressing the more complex task, which consists in assigning the congruent meaning to the lexical items employed inside the glosses. This will be the theme of a further article, discussing how rule-based procedures can be employed in typical Word Sense Disambiguation tasks.

## X. References

[1] Ahlswede T. and Evens M.W. 1988. Generating a Relational Lexicon from a Machine Readable Dictionary. *International Joural of Lexicography*, 1.3: 214-237.

[2] Artale A., Magnini B., Strapparava C., 1997. Lexical Discrimination with the Italian Version of Wordnet.

[3] Baker C.F., Fillmore Ch.J, Cronin B., 2003. The Structure of the FrameNet Database, *International Journal Lexicography* 16 (3): 281-296.

[4] Barnbrook G., Danielsson P., Mahlberg,M. (eds.) 2004 *Meaningful Texts: The Extraction of Semantic Information from Monolingual and Multilingual Corpora*, London - New York: Continuum International Publishing Group Ltd.

[5] Baroni M. and Zamparelli R, 2010. *N*ouns are vectors, adjectives are matrices: Representing adjective-noun constructions in semantic space. *Proceedings of the Conference on Empirical Methods in Natural Language Processing,* 2010, East Stroudsburg PA: ACL, 1183-1193

[6] Bentivogli L., Girardi C., Pianta E., 2003., The MEANING Italian Corpus. *Corpus Linguistics 2003 Conference, 2003,* pp. 103-112. (Corpus Linguistics 2003 Conference, Lancaster, United Kingdom)

[7] Bierwisch, M., 1967. Some semantic universals of German adjectivals. *Foundations of Language* 3, pp. 1-36.

[8] Bindi R., Calzolari N., Monachini N., Pirelli V., Zampolli A., 1994. *C*orpora and computational lexica. Integration of different methodologies of lexical knowledge acquisiton. *Literary and Linguistic Computing*, 9, 1 (1994), 29-46.

[9] Boas, H.C., 2009. 'A frame-semantic approach to identifying syntactically relevant elements of meaning'. In P. Steiner, H.C. Boas and S. Schierholz (eds.), *Contrastive Studies and Valency. Studies in Honour of Hans Ulrich Boas*. Frnkfurt/New York, Peter Lang 119-149.

[10] Bosco C. and Mazzei A. 2012. *The Evalita 2011 Parsing Task: the Dependency Track.*

[11] Byrd R.J., Calzolari N., Chodrow S., Klavans J., Neff M.S. and Rizk O.A. 1984. Tools And Methods For Computational Lexicology. "*Computational Linguistics*", Volume 13, Numbers 3-4.

[12] Calzolari N. (1984) Detecting Patterns in a Lexical Database, *Proceedings in the 10th International Conference on Computational Linguistics. COLING'84*. Stanford, California, 170-173.

[13] Calzolari N. (1991). 'Acquiring and Representing Semantic Information in a Lexical Knowledge Database'. In Pustejovsky J. and Bergler S. (eds.) *Proceedings of the First SINGLEX Worshop on Lexical Semantics and Knowledge Representation*. London, Springer: 235-243.

[14] Dolan W. B., Vanderwende L and Richardso S.D. 2000, 'Polysemy in a Broad Coverage Natural Processing System'. In Ravin Y. and Leacock C. (eds), *Polysemy. Theoretical and Computational Approaches.* Oxford, Oxford University Press.: 178-204.

[15] De Mauro, T., 1968. *Per una teoria formalizzata del noema lessicale* in De Mauro 1971: 115-160; (also in De Mauro 1989: 235-282).

[16] Fellbaum C., Delfs L., Wolff S., Palmer M., 2004. Word meaning in dictionaries, corpora and the speaker's mind. In Barnbrook, Danielsson, Mahlberg (eds.), *Meaningful Texts: The Extraction of Semantic Information from Monolingual and Multilingual Corpora*, London - New York: Continuum International Publishing Group Ltd.: 31-38.

[17] Fontenelle T., 1997. Using a Bilingual Dictionary to Create Semantic Networks. *International Journal of Lexicography*, 10.4: 275-303.

[18] Fontenelle T. 2008. *Practical Lexicography.* Oxford University Press, New York.

[19] Granger S., Paquot M., 2012. *Electronic Lexicography.* Oxford University Press.

[20] Greimas, A.J., 1996. *Del senso* [*Du sens*, Paris, 1970], Milano, Bompiani

[21] Grella, M., Nicola, M., Christen D., 2011. Experiments with a Constraint-based Dependency Parser. *Evalita 2011 Dependency Parsing Task*. (http://www.evalita.it/sites/evalita.fbk.eu/files/working_notes2011/Parsing/DEP_PARS_PARSIT.pdf)

[22] Horskotte, G., 1982. *Sprachliches Wissen: Lexikon oder Enzyklopädie?,* Bern-Stuttgart-Wien, Verlag Hans Huber

[23] Jackendoff, R., *Semantics and Cognition.* MIT Press, Cambridge, Mass.. 1986).

[24] Jackendoff R., (1999). *Semantic Structures*. MIT Press, Cambridge.

[25] *Johnson-Laird Ph. N., (1983), Mental Models.* Cambridge Univ. Press. Cambridge.

[26] Labov, W., 1977. *Il continuo e il discreto nel linguaggio*, Bologna, Il Mulino.

[27] Lesk, M., 1986. *Automated Sense Disambiguation Using Machine-readable Dictionaries: How to Tell a Pine Cone from an Ice Cream Cone*. Proceedings of the 1986 SIGDOC Conference.

[28] Magnini B., Strapparava C., (1994), *Costruzione di una base di conoscenza lessicale per l'italiano basata su ItalWordNet*, Atti del XXVIII Congresso della Società di Linguistica Italiana, Palermo, 415-423

[29] Magnini B., Strapparava C., Pezzulo C., GhiozzoA., 2003. *The Role of Domain Information in Word Sense Disambiguation*. Journal of Natural Language Engineering (on Sensval-2), 9.

[30] Markowitz, J., Ahlswede, T., Evens, M., 1986. *Semantically significant patterns in dictionary definitions*. ACL Conference Proceedings, 112-119.

[31] McCarthy D. and Carroll J., 2003. Disambiguating nouns, verbs, and adjectives using automatically acquired selectional preferences. *Computational Linguistics*, 29(4):639–654, December.

[32] McCawley, J.D., 1968. *The role of semantics in a grammar*. In *Universals in Linguistic Theory*, E. Bach and R. Harms (eds.), 124-169. New York: Holt, Rinehart.

[33] Mel'čuk, I.A., 2012. *Semantics. From Meaning to Text*.

[34] Miller G.A. 1995. Wordnet: a lexical database for english. *Communications of the ACM*, 38(11):39–41.

[35] Nerlich B., Todd Z., Vimala H. (eds.), 2003. *Polysemy: Flexible Patterns of Meaning in Mind and Language*, Walter de Gruyter

[36] Peters W. and Kilgarriff A., 2000. Discovering Semantic Regularity in Lexical Resources. *International Journal of Lexicography*, 25.2: 152-190.

[37] Prandi M., 2004 *The Building Blocks of Meaning*. Ideas for a Philosophical Grammar. John Benjamins ed. Amsterdam/Philadelphia.

[38] Prieto, L.J., 1967. *Principi di noologia. Fondamenti della teoria funzionale del significato*, Roma, Ubaldini

[39] Prieto, L., 1976. *Pertinenza e pratica. Saggio di semiotica*, Milano, Feltrinelli

[40] Pustejovsky J., 1995. *The Generative Lexicon.* MIT Press, Cambridge

[41] Pustejovsky J. (1991). *The Generative Lexicon*, "Computational Linguistics", 17: 71-77

[42] Pustejovsky J., Bergler S., 1992. *Lexical semantics and knowledge representation.* First SIGLEX Workshop, Berkeley, CA, USA, June 1991. Springer

[43] Pustejovsky J., (ed.), 1993. *Semantics and the Lexicon*. MIT Press, Cambridge, MA.

[44] *Pustejovsky J., 1995. The Generative Lexicon.* MIT Press, Cambridge

[45] Searle, J.R., 1994. *La riscoperta della mente*, Torino, Bollati Boringhieri.

[46] Resnik, P. (1997) Selectional Preference and Sense Disambiguation, *In Proceedings of ACL Siglex Workshop on Tagging Text with Lexical Semantics, Why, What and How?*, Washington, 1997.

[47] Sahlgrem M., 2008. *The Distributional Hypothesis. From context to meaning: Distributional models of the lexicon in linguistics and cognitive science*. Rivista di Linguistica, 20,1: 33-53 .

[48] Sinclair J., Ball J. (1996). *EAGLES Preliminary Recommendations onText Typology.* (url).

[49] Véronis J., Ide. N., 1995. *Large Neural Networks for the Resolution of Lexical Ambiguity*. In *Computational Lexical Semantics* (P. Saint-Dizier ed.) .Cambridge University Press, 1995.

[50] Vossen P., Meijs W. and Den Broeder M., 1989. Meaning and Structure in Dictionary Definitions. In Boguraev B. and Briscoe T. (eds.) Computational Lexicography for Natural Language Processing. London, Longman: 171-192.

[51] Vossen P., 1992. *The automatic construction of a knowledge base from dictionaries: a combination of techniques*, in: H. Tommola, K. Tarantola, T. Salmin Tolonen, J. Schopp (eds.) *Proceedings of the 5th Euralex International Congress on Lexicography*, Tampere, Finland, 199: 311-326..

[52] Vossen P., et. al. 1998, *The EuroWordNet base concepts and top ontology.* Final Version (http://www.researchgate.net/)

[53] Wilks, Y., Charniak, E. (eds)., 1976 *Computational Semantics. An Introduction to Artificial Intelligence and Natural Language Understanding*. Amsterdam: North-Holland

[54] Wilks, Y. (1977). Good and bad arguments about semantic primitives. *Communication & Cognition* 10(3/4), pp. 181-221.
[55] Wilks Y. and Mark Stevenson M., *The grammar of sense: using part-of speech tags as a first step in semantic disambiguation*, Natural Language Engineering **4** (1998), no. 2, 135–143
[56] Wilks, Y., D. Fass, C. Guo, J. MacDowland, T. Plate, B. Slator, 1990. *Providing Machine Tractable Dictionary Tools*. In J. Pustejovsky (ed.), *Semantics and the Lexicon*. MIT Press, Cambridge, MA.
[57] Wilks, Y., Slator, B., Guthrie, L. (1996) *Electric Words: dictionaries, computers and meanings*. Cambridge, MA: MIT Press.
[58] Wilks, Y., Brewster, C., 2009. *Natural Language Processing as a Foundation of the Semantic Web*. Now Press: London
[59] Zampolli A., Cappelli A., 1981. (eds.), *The possibilities and limits of the computer in producing and publishing dictionaries*. Proceedings of the European Science Foundation Workshop, Pisa, 1981, Istituto di Linguistica Computazionale - Giardini, Roma - Pisa, 1984, pp. 77-82. (Linguistica Computazionale, 3, 1983).
[60] Zernik U. (ed.). 1991. *Lexical Acquisition: Exploring On Line Resources to build a Lexicon*. Hillsdale, Lawrence Erlbaum Ass.
[61] Violi, P., 1997. *Significato ed esperienza,* Milano, Bompiani.
[62] Wittgenstein, L., 1974. *Ricerche filosofiche,* Torino, Einaudi.

# XI. ANNEXES

TABLE 1. Word's frequency in syntactic-head position

| Word | N. of occurrences as syntactical head of a gloss | |
|---|---|---|
| ciò | *pron. followed by a relative clause* | 7087 |
| chi | *person-specific relative pronoun* | 1717 |
| essere | *to be* | 1264 |
| persona | *person* | 1219 |
| fare | *to do; mostly causative: far cadere* | 1119 |
| parte | *part (of)* | 834 |
| relativo | *belongs to* | 743 |
| insieme | *set (of)* | 742 |
| avere | *to have* | 622 |
| rendere | *to make* | 582 |
| in modo+ADJ | *in a + ADJ+ way* | 571 |
| mettere | *to place, to put* | 430 |
| elemento | *element, element of* | 416 |
| diventare | *to become* | 403 |
| privare | *to deprive* | 397 |
| pianta | *plant* | 388 |
| atto | *act/action of* | 377 |
| quello | *demonstrative* | 342 |
| azione | *action (which consists in...)* | 339 |
| dare | *to give (abstract sense as well)* | 335 |
| strumento | *instrument, tool* | 334 |
| quella | *demonstrative* | 331 |
| luogo | *location, place* | 324 |
| complesso | *complex, set* | 315 |
| cosa | *thing* | 310 |
| usare | *to use (for)* | 271 |
| potere | *can (modal)* | 260 |
| operazione | *operation, complex action* | 259 |
| prendere | *to take, to assume (also related to quality or manner)* | 255 |
| nome | *name (of)* | 254 |
| sostanza | *substance, matter* | 251 |
| quantità | *quantity (of)* | 246 |
| tipo | *type (of)* | 244 |
| denominazione | *denomination (of)* | 240 |
| dire (detto di) | *denomination (of)* | 226 |
| condizione | *condition, state* | 225 |
| dispositivo | *instrument, tool* | 224 |
| movimento | *movement* | 218 |

TABLE 2. Dependency tree of the gloss "gorilla"

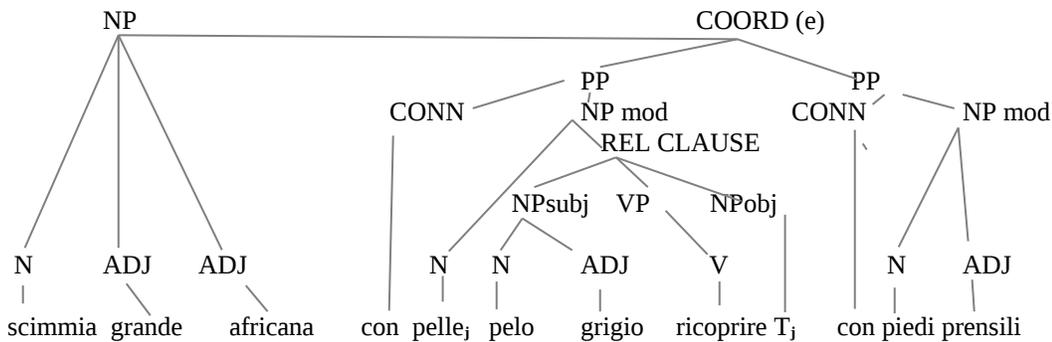